# Automated scoring of the Ambiguous Intentions Hostility Questionnaire

## using fine-tuned large language models

Yizhou Lyu*[1], Dennis Combs[2], Dawn Neumann[3], Yuan Chang Leong[*4,5,6]

[1] Department of Psychology, University of California, Los Angeles, Los Angeles, CA 90025

[2] Psychology & Counseling, University of Texas at Tyler, Tyler, TX 75799

[3] Physical Medicine and Rehabilitation, JFK Johnson Rehabilitation Institute, Edison, NJ 08820

[4] Department of Psychology, University of Chicago, Chicago, IL 60637

[5] Institute of Mind and Biology, University of Chicago, Chicago, IL 60637

[6] Neuroscience Institute, University of Chicago, Chicago, IL 60637

**Author Note**

We have no known conflict of interest. This research was supported by computational resources provided by the University of Chicago Research Computing Center and the University of Chicago Data Science Institute. Some of this work was supported by the National Institute on Disability, Independent Living, and Rehabilitation Research (NIDILRR), award # 90IF00-95-01-00.

* Correspondence should be addressed to Yizhou Lyu at lyulouisa1@ucla.edu and Yuan Chang Leong at ycleong@uchicago.edu.

**Abstract**

Hostile attribution bias is the tendency to interpret social interactions as intentionally hostile. The Ambiguous Intentions Hostility Questionnaire (AIHQ) is commonly used to measure hostile attribution bias, and includes open-ended questions where participants describe the perceived intentions behind a negative social situation and how they would respond. While these questions provide insights into the contents of hostile attributions, they require time-intensive scoring by human raters. In this study, we assessed whether large language models can automate the scoring of AIHQ open-ended responses. We used a previously collected dataset in which individuals with traumatic brain injury (TBI) and healthy controls (HC) completed the AIHQ and had their open-ended responses rated by trained human raters. We used half of these responses to fine-tune the two models on human-generated ratings, and tested the fine-tuned models on the remaining half of AIHQ responses. Results showed that model-generated ratings aligned with human ratings for both attributions of hostility and aggression responses, with fine-tuned models showing higher alignment. This alignment was consistent across ambiguous, intentional, and accidental scenario types, and replicated previous findings on group differences in attributions of hostility and aggression responses between TBI and HC groups. The fine-tuned models also generalized well to an independent nonclinical dataset. To support broader adoption, we provide an accessible scoring interface that includes both local and cloud-based options. Together, our findings suggest that large language models can streamline AIHQ scoring in both research and clinical contexts, revealing their potential to facilitate psychological assessments across different populations.

## Introduction

Imagine a scenario where a driver is abruptly cut off in traffic. While some might dismiss it as a careless oversight, others may perceive it as a deliberate act of aggression. This tendency to interpret ambiguous situations as intentionally hostile is often referred to as hostile attribution bias (Dodge et al., 2015; Epps & Kendall, 1995; Nasby et al., 1980), and is a well-documented predictor of aggressive behavior and interpersonal conflict (Klein Tuente et al., 2019; Pettit et al., 2010). For example, a driver with hostile attribution bias might respond to being cut off by aggressively tailgating the other driver, escalating a minor incident into a potentially dangerous situation. Hostile attribution bias is not only associated with aggressive behavior but is also linked to a range of psychiatric disorders, including anxiety, depression, and schizophrenia (An et al., 2010; Bailey & Ostrov, 2008; Eysenck et al., 1991; H. L. Smith et al., 2016). People with traumatic brain injury (TBI) are also susceptible to hostile attribution bias and related anger and aggression (Neumann et al., 2017, 2017, 2020; Neumann, Sander, Witwer, et al., 2021).

Individuals who consistently interpret others' actions as hostile may be more prone to developing anxiety and depressive disorders due to the constant perception of threat and conflict in their environment ((Mathews & Macleod, 1985; Mogg et al., 2006). Moreover, this bias has been implicated in the exacerbation of symptoms in individuals with schizophrenia, where misinterpretations of social cues can lead to heightened paranoia and social withdrawal (An et al., 2010; Buck et al., 2023). Given its wide-ranging impact on antisocial behavior, interpersonal relationships and mental health outcomes, studying hostile attribution bias is crucial for developing interventions that can reduce its harmful effects and improve individual and societal well-being.

**Measuring Hostile Attribution Bias**

Existing measurements of hostile attribution bias can be broadly categorized into three types: personality inventories that assess generalized tendencies towards perceiving hostility, tasks involving responses to simplified social stimuli that isolate specific aspects of social perception, and assessments based on reactions to complex social situations that closely mimic real-world interactions (see Wagels & Hernandez-Pena, 2024 for review).

### Personality Inventories

Personality inventories are designed to evaluate an individual's predisposition to perceive hostility in various situations by relying on self-reported data. These inventories typically consist of questionnaires where individuals are asked to rate how frequently they experience thoughts or feelings that suggest a hostile interpretation of others' behaviors. An example is the Cook-Medley Hostility Scale (Barefoot et al., 1989; Cook & Medley, 1954), which consists of 50 true-false statements related to dispositional aggression, cynicism and the tendency to attribute hostile intentions. The Cook-Medley Hostility Scale has been shown to correlate with increased self-reported anger, overt hostile behavior, and a tendency to blame others for disagreements during conflict (T. W. Smith et al., 1990). While such personality inventories are effective in capturing broad tendencies and dispositional traits, they rely on individuals' metacognitive ability to assess and report their own tendencies, which may not always be reliable (Paulhus, Delroy L & Vazire, Simine, 2007). Additionally, because these inventories do not incorporate the context in which social interactions occur, they may not accurately reflect how these tendencies manifest in real-world situations.

### Judgments of Ambiguous Social Cues

A second approach to measuring hostile attribution bias involves tasks that present participants with simplified, decontextualized social stimuli, such as ambiguous facial expressions (Gennady G Knyazev et al., 2009; Schönenberg & Jusyte, 2014) or short clips of people laughing (Ethofer et al., 2020; Martinelli et al., 2019; Plate et al., 2022). Participants are then asked to interpret the social intent behind these stimuli, typically by indicating whether they perceive the social cue as friendly, neutral, or hostile. These tasks are designed to assess the automatic cognitive biases that may lead to hostile attributions in everyday interactions. One of the strengths of this approach is that it allows researchers to focus on how individuals interpret specific social cues in a controlled environment, which can help in identifying the fundamental cognitive processes underlying hostile attribution bias. However, similar to personality inventories, these tasks lack information about the social context and may not fully capture how social cues and contextual information are combined in real-world interactions.

### Judgments of Ambiguous Social Situations

The third approach to measuring hostile attribution bias involves presenting participants with complex, context-rich social scenarios that resemble real-life interactions. These scenarios are conveyed through written vignettes (Gagnon et al., 2015; Neumann et al., 2020), audio recordings (Lyu et al., 2024; Neumann, Sander, Witwer, et al., 2021), video clips (Coccaro et al., 2021, 2022), or virtual reality devices (Hummer et al., 2023). Participants are then asked to interpret the intentions behind the actions of others within these scenarios. This approach provides a more ecologically valid assessment of hostile attribution bias, as it incorporates the social context that is critical for interpreting others' actions. Tools such as the Ambiguous Intentions Hostility Questionnaire (AIHQ; Combs et al., 2007), Epps Scenarios (Epps & Kendall, 1995), the Social Information Processing-Attribution and Emotional Response Questionnaire (SIP-AEQ; Coccaro et al., 2009), and the Video Social-Emotional Information Processing assessment (V-SEIP; Coccaro et al., 2017) are examples of measures that use this method.

These measures vary in the types of responses they collect from participants. Most measures include behavioral ratings, where participants are asked to evaluate aspects of social interactions, such as the perceived hostility of others' intentions, on a numerical scale. To allow for a more detailed characterization of how participants interpreted a situation, measures such as the V-SEIP and SIP-AEQ also ask participants to indicate the extent to which they agree with a predefined set of benign or hostile attributions. This approach is advantageous because it provides quantitative data that is easy to analyze and compare across participants.

In addition to structured response formats, some measures, such as the Epps Scenarios and the AIHQ, incorporate open-ended responses where participants describe their interpretations of the situations and how they would respond. This approach offers several advantages over relying solely on self-reported ratings or predefined statements. First, the open-ended responses focus on the content of participants' interpretations, providing insight into the underlying thought processes that drive their perceptions of hostility. Furthermore, when participants articulate their thoughts in their own words, it can reveal subtleties in language use, tone, or emphasis that would be missed when participants provide numerical ratings or respond to predefined statements. Finally, numerical scales are prone to variability in how individuals interpret scale points, which introduces noise into the data. Open-ended responses, when scored by trained raters using standardized criteria, allow for greater consistency and comparability across participants.

### *The Ambiguous Intentions Hostility Questionnaire*

The AIHQ is one of the most widely used tools for assessing hostile attribution bias in research and clinical settings (see B. Buck et al., 2023 for recent review; Combs et al., 2007). Using the AIHQ, researchers have consistently documented elevated levels of hostile attribution bias in various clinical populations, including in individuals with schizophrenia and traumatic brain injury (Buck et al., 2017; Healey et al., 2015; Neumann et al., 2020). This heightened bias has been shown to correlate with paranoia, anxiety, depression, anger and aggression (Combs et al., 2009; Darrell-Berry et al., 2017; Jeon et al., 2013; Mancuso et al., 2011; Neumann et al., 2017, 2020; Neumann, Sander, Perkins, et al., 2021; Neumann, Sander, Witwer, et al., 2021), and to predict negative life outcomes such as increased interpersonal conflict and lower quality of life (Hasson-Ohayon et al., 2017; Waldheter et al., 2005). Additionally, the AIHQ is frequently employed to evaluate the efficacy of interventions aimed at reducing hostile attribution bias (Lahera et al., 2013; Neumann et al., 2023), making it a valuable tool for both understanding and addressing the cognitive distortions that underlie hostility in social interactions.

The AIHQ consists of 15 short vignettes that describe negative social situations with varying levels of intentionality, including intentional, accidental, and ambiguous scenarios. For example, one vignette describes the participant walking past a group of teenagers and hearing them laugh. An individual with high hostile attribution bias may interpret the laughter as mocking or directed at them, while an individual with low hostile attribution bias may view it as unrelated or stemming from a benign interaction among the teenagers. Participants are instructed to read these vignettes and imagine the situations happening to them. They then respond to a series of open-ended and structured prompts that guide them to describe the perceived intent behind the actions, their emotional reactions, and how they would respond behaviorally. Open-ended responses are used to capture participants' spontaneous interpretations and provide qualitative data that can be scored using standardized criteria. Additionally, participants rate the perceived hostility of the other person's intent and the degree of blame assigned to the individual on numerical scales, allowing for quantitative analysis.

While the open-ended responses in the AIHQ provide rich and nuanced insights into participants' thought processes, scoring these responses presents significant challenges. Accurate scoring requires trained raters who can evaluate responses using standardized criteria, a process that demands substantial time and resources for training and implementation. Even with well-trained raters, variability between scorers can arise. These challenges limit the widespread adoption of the AIHQ, particularly in clinical and research settings where resources and time are often constrained. Developing automated scoring methods for open-ended responses could address these limitations by providing an objective, consistent, and scalable approach to analyzing responses. Such advancements would make the AIHQ more appealing and practical for clinicians and researchers, enabling broader application and facilitating large-scale studies that assess hostile attribution bias using the AIHQ.

**Automated Scoring Methods**

There have been several computational methods developed for automated scoring of open-ended responses for emotional content. Most of these methods are dictionary-based approaches that use predefined word lists. For example, the Dictionary of Affect in Language (DAL) assigns scores to words based on their normed pleasantness and activation level, and computes the mean rating of scored words along these dimensions (Mossholder et al., 1995; Whissell, 2009). Similarly, the Linguistic Inquiry and Word Count (LIWC) program categorizes words into psychological, emotional, and linguistic categories based on a pre-defined dictionary (Pennebaker et al., 2015). LIWC calculates the proportion of words in a text that fall within each category as a measure of the extent to which the text reflects specific psychological or emotional states. For instance, a high proportion of words categorized as "negative affect" (e.g., "sad," "angry") would indicate a negative emotional tone in the response.

DAL, LIWC, and related dictionary-based methods provide an efficient and consistent means of analyzing large volumes of text data. However, these approaches are not well-suited for scoring responses from the AIHQ. Firstly, they are not specifically designed to assess attribution of hostility or aggression, which are the central constructs measured by the AIHQ. Furthermore, dictionary-based methods are inherently constrained by the predefined word lists they rely upon, which may fail to capture relevant words or phrases that are not included in the dictionary. This limitation reduces their sensitivity to nuances in participant responses. Moreover, these methods lack the ability to disambiguate context, a critical factor in interpreting the underlying intent or emotional tone of a response. This shortcoming is especially problematic for the AIHQ, where responses are often brief, requiring contextual understanding of the situation described in the vignette to evaluate the level of hostility or aggression accurately. Thus, advanced methods capable of interpreting nuanced language and context-specific cues are needed for automated scoring of the open-ended questions on the AIHQ.

### Language Analysis using Large Language Models

The development of large language models (LLMs) offers promising new methods for analyzing open-ended responses in psychometric assessments. LLMs such as GPT (Brown et al., 2020) and Flan (Wei et al., 2022) are trained on extensive text corpora, allowing them to learn patterns of language use, grammar, and semantic meaning through next-word prediction. Additionally, instruction tuning and reinforcement learning with human feedback enhance the models' ability to provide accurate and contextually appropriate analyses, further improving their capacity to capture meaning and intent within text. To that end, they have been shown to excel in processing and generating human-like text, as illustrated by their ability to respond to complex linguistic inputs in a manner that closely mirrors human language use (Ye et al., 2023).

LLMs have demonstrated significant potential for use in automated psychological text analysis (Demszky et al., 2023), producing outputs that align closely with human ratings across a variety of psychological constructs, including valence and arousal (Rathje et al., 2023; Yang et al., 2024), offensive language and insults (Rathje et al., 2023), and creativity (DiStefano et al., 2024). Unlike dictionary-based methods, which rely on static word lists, LLMs are highly adaptable and capable of interpreting the context-dependent nuances of brief and complex participant responses. This capability makes them well-suited for tasks such as assessing attribution of hostility, blame, and emotional tone in open-ended responses.

One of the most powerful techniques for enhancing the performance of LLMs in specific applications is fine-tuning. Fine-tuning involves training a pre-existing language model on a domain-specific dataset, allowing it to specialize in tasks or subject areas beyond its general training (Howard & Ruder, 2018). By exposing the model to examples that are representative of the task at hand, fine-tuning refines the model's parameters, enabling it to better understand and predict task-relevant patterns in the data. When fine-tuned on datasets with ratings provided by trained human raters, the model learns to align its predictions with human judgment by incorporating the linguistic and contextual features that human raters consider important.

**The Present Study**

The goal of the present study was to develop and validate LLM-based automated scoring methods for the open-ended questions in the Ambiguous Intentions Hostility Questionnaire (AIHQ). Specifically, we evaluated the performance of two state-of-the-art language models, GPT-3.5-Turbo and Flan-T5-Large, in predicting attribution of hostility and aggression response scores based on participants' text responses in a dataset comprised of manually scored AIHQ responses from participants with traumatic brain injury (TBI) and healthy controls (Neumann et al., 2020; see more detail below). The dataset was divided into two halves: one half was used to fine-tune the language models to predict scores assigned by human raters, while the other half was held-out for testing the models' performance. We then assessed the model accuracy by correlating the model-predicted scores with scores provided by trained human raters. In addition to evaluating the accuracy of the model predictions, we tested whether the automated scoring methods could replicate previously documented group differences between TBI patients and healthy controls. Finally, to assess the generalizability of the models to a different population, we applied the fine-tuned models to a separate dataset of AIHQ responses collected from undergraduate students (Combs et al., 2013).

## Methods

### Ambiguous Intentions Hostility Questionnaire (AIHQ)

The AIHQ consists of 15 hypothetical scenarios designed to evaluate participants' perceptions and responses to social situations. These scenarios are categorized into three types based on the nature of the described behaviors: (1) intentional actions, which are clearly deliberate, (2) ambiguous actions, where the intent is unclear, and (3) accidental actions (n = 5 per category). Following each vignette, participants rated how angry they felt and how much they blamed the character on a 5-point scale, and the perceived intentionality of the action on a 6-point scale. They also answered two open-ended questions that measure (1) *attribution of hostility*, asking why the character in the vignette acted the way they did, and (2) *aggression response*, asking how the participant would respond in that situation.

### Dataset 1 (Neumann et al., 2020)

The first dataset used in this study was drawn from a previous investigation assessing negative attributions in participants with traumatic brain injury (TBI) and healthy controls (HCs). The study included 85 adults with TBI and 85 HCs, recruited from two rehabilitation hospitals in Indiana and Texas, and was approved by the ethics review boards at both locations. All participants provided consent prior to participation. The TBI group consisted of individuals with injury severities ranging from complicated mild to severe, with a Glasgow Coma Score of < 13 at the time of injury, posttraumatic amnesia ≥ 24 hours, loss of consciousness ≥ 30 minutes, or CT scan showing intracranial abnormality. Participants were a minimum of six months post-injury. Both TBI and HC participants were free from other neurological disorders, major psychiatric conditions affecting social cognition, and developmental disabilities. All were 18 years or older, spoke fluent English, and had adequate expressive language and comprehension abilities as determined by the Discourse Comprehension Test.

The AIHQ was administered as part of a broader battery of assessments. The open-ended questions were scored by two trained raters (intraclass correlation coefficients: *attribution of hostility* = 0.80, *aggression response* = 0.73). Scale scores for each of the three scenario types were calculated by averaging the five ratings within each scenario type, and then an overall mean score for each scale was derived across all scenario types. The dataset was randomly divided into two groups, each containing an equal number of TBI and HC participants (Table 1; Figure 1). One group was used for fine-tuning the automated scoring models, while the other was reserved for testing the models' performance and validating their ability to generalize across participant responses.

**Table 1.** *Participant demographics (Dataset 1)*

| Variables | Training dataset | | Testing dataset | |
|---|---|---|---|---|
| | **TBI subjects (n = 42)** | **HC subjects (n = 42)** | **TBI subjects (n = 43)** | **HC subjects (n = 43)** |
| Age, mean (SD) | 40.4 (14.2) | 39.9 (15.8) | 39.7 (12.6) | 41.2 (14.6) |
| Sex, *n* (%) | | | | |
| Male | 22 (52.4%) | 20 (47.6%) | 25 (58.1%) | 23 (53.5%) |
| Female | 20 (47.6%) | 22 (52.4%) | 18 (41.9%) | 20 (46.5%) |
| Race, *n* (%) | | | | |
| White | 32 (76.2%) | 33 (78.6%) | 33 (76.7%) | 31 (72.1%) |
| Asian | 0 (0%) | 0 (0%) | 1 (2.32%) | 0 (0%) |
| Black | 9 (21.4%) | 8 (19.0%) | 8 (18.6%) | 11 (25.6%) |
| Native American | 1 (2.38%) | 1 (2.38%) | 0 (0%) | 0 (0%) |
| Other | 0 (0%) | 0 (0%) | 2 (4.65%) | 1 (2.33%) |

**Dataset 2 (Combs et al., 2013)**

The second dataset for model validation was drawn from a previous study in a non-clinical undergraduate sample (Table 2). The sample included 146 undergraduate students who completed measures of paranoia, emotion perception, attributional style, and social functioning in a single 2-hour session. All participants received extra credit in their undergraduate class for participation in the study.

The Ambiguous Intentions Hostility Questionnaire (AIHQ) was administered as part of the attributional style assessment. Responses to the two open-ended questions were independently scored by two research assistants blinded to the study. These assistants were trained on rating AIHQ, including ratings of sample responses and feedback sessions to achieve intraclass correlation coefficients (ICCs) of at least 0.80 with a criterion-trained rater. The full AIHQ dataset of 146 participants was used to validate the trained models from Dataset 1, testing the trained models' effectiveness in scoring AIHQ in a different nonclinical group.

**Table 2.** *Participant demographics (Dataset 2)*

| N | 144 |
|---|---|
| Age, mean (SD) | 22.5 (5.84) |
| Sex, *n* (%) | |
|     Male | 27 (18.75%) |
|     Female | 117 (81.25%) |
| Race, *n* (%) | |
|     White | 103 (73.05%) |

*Note.* Detailed breakdown of race information was not available

**Model Fine-tuning**

We used two large language models for automated scoring of the AIHQ responses: GPT-3.5-Turbo and Flan-T5-Large. GPT-3.5-Turbo is a large language model developed by OpenAI, optimized for conversational tasks. Flan-T5-Large is a large language model published by Google, designed to enhance zero-shot learning through instruction tuning (Wei et al., 2022). To enhance GPT-3.5-Turbo's reliability for generating single numeric ratings, we set the temperature parameter to 0, ensuring deterministic outputs that reflect the most probable response. Additionally, we set the maximum token limit to 10 to help prevent extraneous text and focus the model on producing a single numeric rating.

To improve large language model's accuracy in providing automated ratings for AIHQ responses, we incorporated supervised fine-tuning and in-context learning. Specifically, we randomly subsampled half of the TBI participants (n = 42) and half of the HC participants (n = 42) in dataset 1 to create a training dataset containing their responses to all 15 AIHQ scenarios (Figure 1). We then reformatted these responses into the fine-tuning formats for each model - a conversational chat format for GPT-3.5-Turbo and a user-question format for Flan-T5-Large (see Additional Information). After fine-tuning, both models were presented with the same *attribution of hostility* and *aggression response* rating prompts to generate automated scores for the other half of participants' responses that were not used in the training dataset.

For Flan-T5-Large, we tracked training loss and ROUGE metrics (ROUGE-1, ROUGE-2, ROUGE-L) across multiple fine-tuning epochs to evaluate the efficacy of the fine-tuning process. A consistent decrease in training and validation loss across epochs, with increasing ROUGE scores, indicates that the model achieved an increasingly better fit to the data. In particular, epoch 3 had both the lowest training and validation loss as well as the highest ROUGE-1, ROUGE-2, ROUGE-L, and ROUGE-LSum values. These results suggest that the third epoch produced the most effective fine-tuned version of Flan-T5-Large, which we selected for subsequent scoring.

**Model Evaluation**

We compared the fine-tuned GPT-3.5-Turbo and Flan-T5-Large models with base versions of the models (i.e., pre-fine-tuned). All four models were used to rate the two open-ended responses in the AIHQ questionnaire from the participants in Dataset 1 who were not used for fine-tuning the model. We evaluated each model's performance by computing the Pearson correlation between the model-generated ratings and the average ratings obtained from human raters. We also compared these correlations to the inter-rater reliability of two human raters (i.e., the correlation between their ratings) to assess if the models were as consistent with human ratings as between the two human raters. In addition, we analyzed model ratings' accuracy separately for ambiguous, intentional, and accidental scenarios and for the two participant groups (TBI vs. HC).

We then assessed whether the TBI group overall received higher ratings for *attribution of hostility* and *aggression response* than the HC group. We also computed correlations between the model ratings on *attribution of hostility*, *aggression response* and participants' self-reported anger, blame, and perceived intentionality ratings to determine whether these relationships replicated those found with human ratings (Neumann et al., 2020). Next, to evaluate the generalizability of the fine-tuned models, we applied the fine-tuned GPT-3.5-Turbo and Flan-T5-Large to a different dataset of AIHQ responses collected from undergraduate participants to examine if they give ratings consistent with human raters' ratings (Combs et al., 2013). We tested the models' performance across the three scenario types (ambiguous, intentional, and accidental) to ensure that ratings were accurate for all three types.

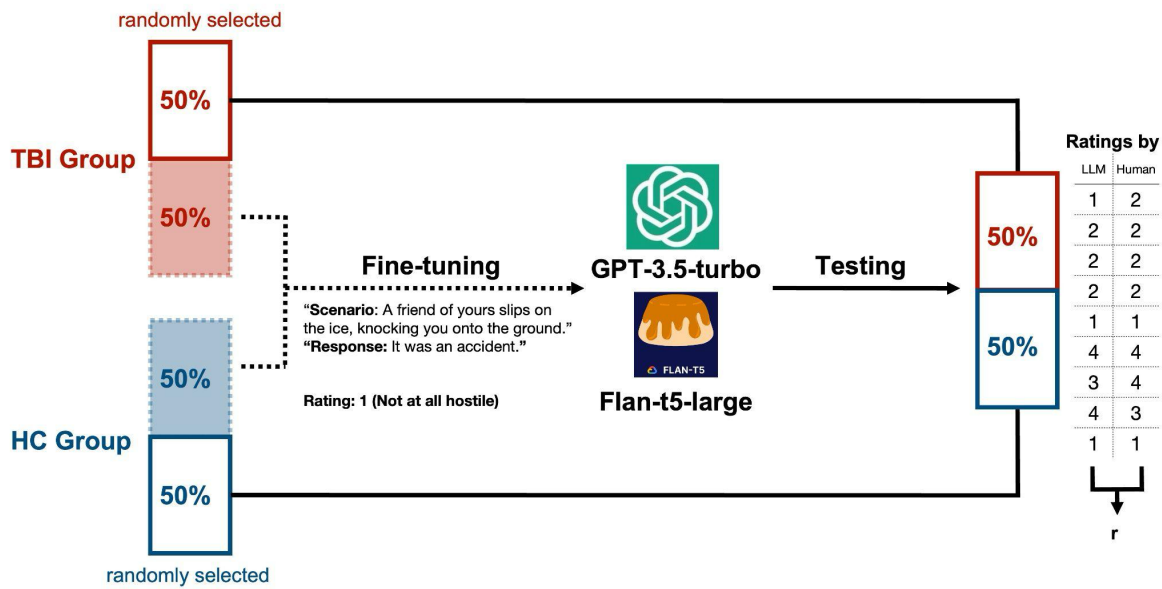

**Figure 1**. **Schematic of model fine-tuning and testing.** We randomly subsampled 50% of TBI and HC data to fine-tune the large language models (LLM; GPT-3.5-turbo and Flan-t5-Large). We then tested the performance of the fine-tuned models on the other 50% of data. Model accuracy was assessed as the correlation between LLM-generated ratings and human ratings.

**Code and data availability**

The pre-trained GPT-3.5-Turbo model is available through the OpenAI API to users with an active and funded account. Due to OpenAI's usage policies, the fine-tuned GPT-3.5-Turbo model cannot be shared. The fine-tuned Flan-T5-Large model can be freely downloaded from the Hugging Face model hub (https://huggingface.co/lyulouisaa/flant5-finetuned-aihqrating). To support automated scoring of AIHQ responses, we provide two user-friendly interfaces. First, a Google Colab notebook allows users to run the Flan-T5-Large model or the base GPT-3.5-Turbo model in a cloud-based environment without requiring local installation. Second, a browser-based interface that runs locally on the user's machine allows researchers to upload AIHQ responses in CSV format and obtain model-generated ratings. Installation instructions, the Google Colab link, and example input templates are available at: https://aihqrating.readthedocs.io/en/latest/index.html.

**Results**

**LLM-generated ratings are correlated with ratings of trained human raters**

The reliability of human-generated ratings served as a benchmark for evaluating the alignment between ratings obtained from LLMs and human raters. The Pearson correlation between the two trained human raters was 0.873 for *attribution of hostility* ratings ($t(84) = 16.94$, $p < .001$) and 0.930 for *aggression response* ratings ($t(84) = 33.90$, $p < .001$). For both *attribution of hostility* and *aggression response*, the ratings generated by the fine-tuned GPT-3.5-Turbo and Flan-T5-Large models closely matched the human ratings, with correlation coefficients exceeding 0.85 (see Table 3; Figure 2). The pre-trained version of GPT-3.5-Turbo was comparable *(attribution of hostility*: $r = 0.840$, $t(84) = 14.21$, $p < .001$; *aggression response*: $r = 0.889$, $t(84) = 17.79$, $p < .001$), but alignment of the pre-trained version of Flan-T5-Large with human ratings was considerably lower (*attribution of hostility*: $r = 0.476$, $t(84) = 4.96$, $p < .001$; *aggression response*: $r = 0.825$, $t(84) = 13.37$, $p < .001$), suggesting the importance of fine-tuning for improving Flan-T5-Large's performance.

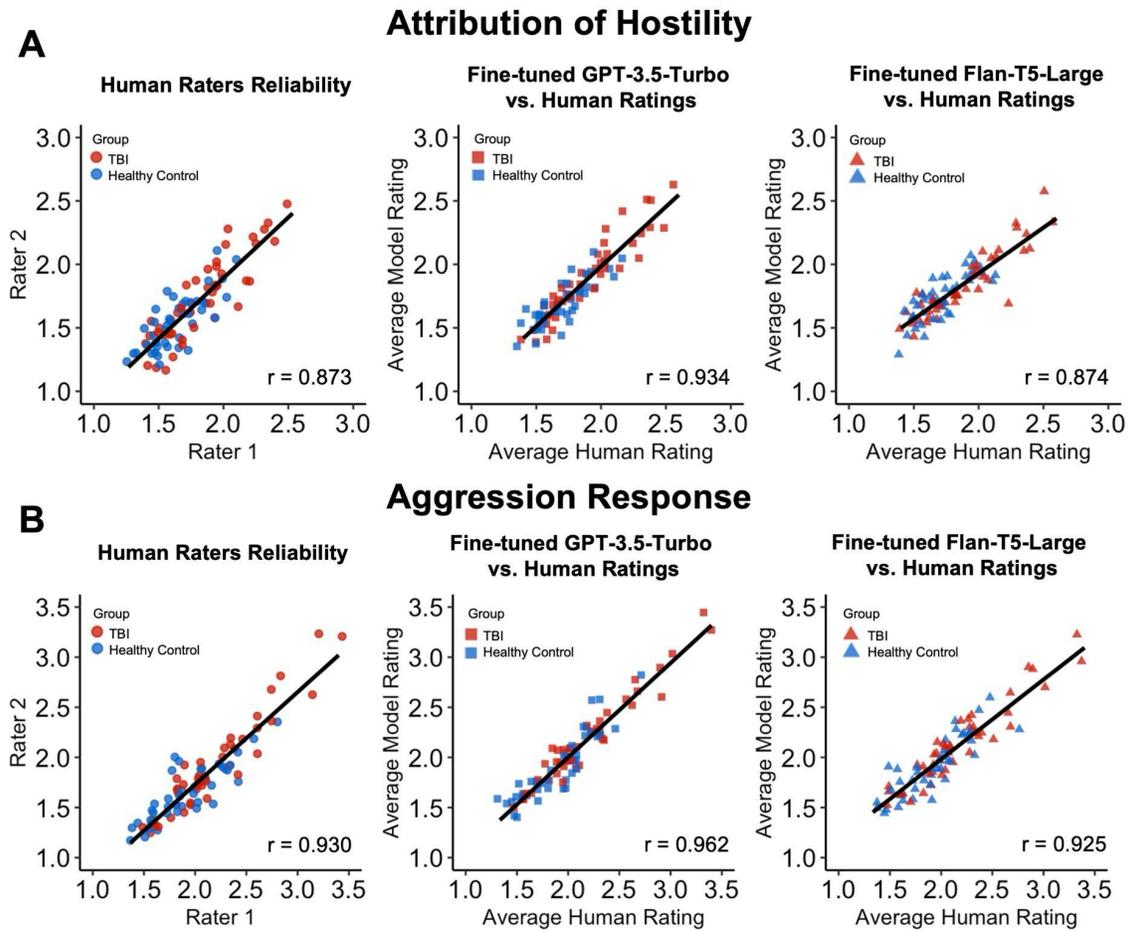

**Figure 2. Ratings given by fine-tuned GPT-3.5-Turbo and Flan-T5-Large correlated with human ratings. A.** *Attribution of hostility:* Each datapoint in the left panel represents one participant's rating for *attribution of hostility* from two human raters. The middle panel compares each participant's average human raters' rating with fine-tuned GPT-3.5-Turbo's rating, and the right panel compares each participant's average human raters' rating with fine-tuned Flan-T5-Large's rating. Red points represent participants from the TBI group, and blue points represent healthy controls. **B.** *Aggression response:* Same as A, but for *aggression response* ratings.

**Table 3.** *Reliability of human-generated ratings and alignment of fine-tuned and pre-trained versions of GPT-3.5-Turbo and Flan-T5-Large with human ratings.*

| | Attribution of Hostility | | Aggression Response | |
|---|---|---|---|---|
| | *r* | *t(df)* | *r* | *t(df)* |
| Human Raters (Inter-rater Reliability) | 0.879 | *t*(84) = 16.94 | 0.929 | *t*(84) = 32.89 |
| GPT-3.5-Turbo (Fine-tuned) | 0.934 | *t*(84) = 23.99 | 0.962 | *t*(84) = 32.28 |
| GPT-3.5-Turbo (pre-trained) | 0.840 | *t*(84) = 14.21 | 0.889 | *t*(84) = 17.79 |
| Flan-T5-Large (Fine-tuned) | 0.874 | *t*(84) = 16.52 | 0.925 | *t*(84) = 22.31 |
| Flan-T5-Large (Pre-trained) | 0.476 | *t*(84) = 4.96 | 0.825 | *t*(84) = 13.37 |

*Note.* All correlations were significant at $p < .001$.

**Language model ratings highly correlate with human ratings across participant groups and scenario types**

Given the superior performance of the fine-tuned models, we focused on the fine-tuned models for subsequent analyses. The model ratings were highly correlated with human ratings across both TBI and HC groups. There were no significant differences between the two groups across the three AIHQ scenario types, indicating that both language models have high validity in rating both TBI and HC participants and all three scenarios in AIHQ. We found similar results with ratings generated by the fine-tuned GPT-3.5-Turbo (Table 4) and those generated by the fine-tuned Flan-T5-Large (Table A1).

**Table 4.** *Correlation between the ratings given by human raters and ratings given by fine-tuned GPT-3.5-Turbo model separately for TBI and HC participants.*

| Variable | Fine-tuned GPT-3.5-Turbo | | Inter-Rater Correlation | |
|---|---|---|---|---|
| | **TBI Group (n = 43)** | **HC Group (n = 43)** | **TBI Group (n = 43)** | **HC Group (n = 43)** |
| **Attributions of hostility** | | | | |
| All scenarios | 0.948 | 0.863 | 0.897 | 0.801 |
| Ambiguous scenarios | 0.887 | 0.835 | 0.916 | 0.802 |
| Intentional Scenarios | 0.882 | 0.849 | 0.758 | 0.730 |
| Accidental scenarios | 0.924 | 0.691 | 0.936 | 0.716 |
| **AIHQ aggression response** | | | | |
| All scenarios | 0.977 | 0.919 | 0.940 | 0.906 |
| Ambiguous scenarios | 0.852 | 0.863 | 0.908 | 0.726 |
| Intentional Scenarios | 0.951 | 0.900 | 0.914 | 0.921 |
| Accidental scenarios | 0.965 | 0.938 | 0.912 | 0.881 |

*Note.* All correlations were significant at $p < .001$.

**Model-generated ratings reproduce differences in *attribution of hostility* and *aggression response* ratings between TBI and HC**

A feature of the AIHQ is its demonstrated sensitivity to differences between clinical and nonclinical populations. We tested whether model-generated ratings replicate previously reported group differences between individuals with TBI and healthy controls (Neumann et al., 2020). The fine-tuned GPT-3.5-Turbo model assigned significantly higher ratings to the TBI group for both attribution of hostility ($t(42) = 3.05$, $p < .01$; Table 5; Figure 3) and anticipated aggressive response ($t(42) = 3.10$, $p < .01$), averaged across all scenario types. This pattern held when the analysis was restricted to ambiguous scenarios, with TBI participants again receiving significantly higher ratings on *attribution of hostility* ($t(42) = 2.73$, $p < .01$) and *aggression response* ($t(42) = 2.91$, $p < 0.01$) toward the scenarios compared to the HC participants. The fine-tuned Flan-T5-Large model yielded comparable results (Table A2), supporting the robustness of these effects across models. Together, these results indicate that the model-generated ratings capture clinically meaningful differences between groups.

**Table 5.** *Testing for the differences in the means of 2 groups for attribution of hostility and aggression response.*

| Variable | Human rated | | | Fine-tuned GPT-3.5-Turbo rated | | |
|---|---|---|---|---|---|---|
| | TBI (n=43) | HC (n=43) | TBI > HC *p* value | TBI (n = 43) | HC (n = 43) | TBI > HC *p* value |
| **Attributions of hostility** | | | | | | |
| All scenarios | 1.90 | 1.71 | .005** | 1.90 | 1.71 | 0.004** |
| Ambiguous scenarios | 1.97 | 1.69 | .017* | 1.98 | 1.68 | 0.009** |
| Intentional Scenarios | 2.42 | 2.35 | .357 | 2.44 | 2.344 | 0.258 |
| Accidental scenarios | 1.29 | 1.11 | .005** | 1.27 | 1.093 | 0.007** |
| **AIHQ aggression response** | | | | | | |
| All scenarios | 2.17 | 1.91 | .005** | 2.17 | 1.91 | 0.003** |
| Ambiguous scenarios | 2.07 | 1.84 | <.001*** | 2.04 | 1.86 | 0.006** |
| Intentional Scenarios | 2.45 | 2.21 | .064 | 2.48 | 2.24 | 0.044* |
| Accidental scenarios | 1.99 | 1.68 | .023* | 1.99 | 1.62 | 0.005** |

*Note.* Average ratings given by large language models trained on 50% of data are significantly higher for the TBI compared to the HC group, reproducing results obtained with human raters. * $p < 0.05$. ** $p < 0.01$, *** $p < 0.001$.

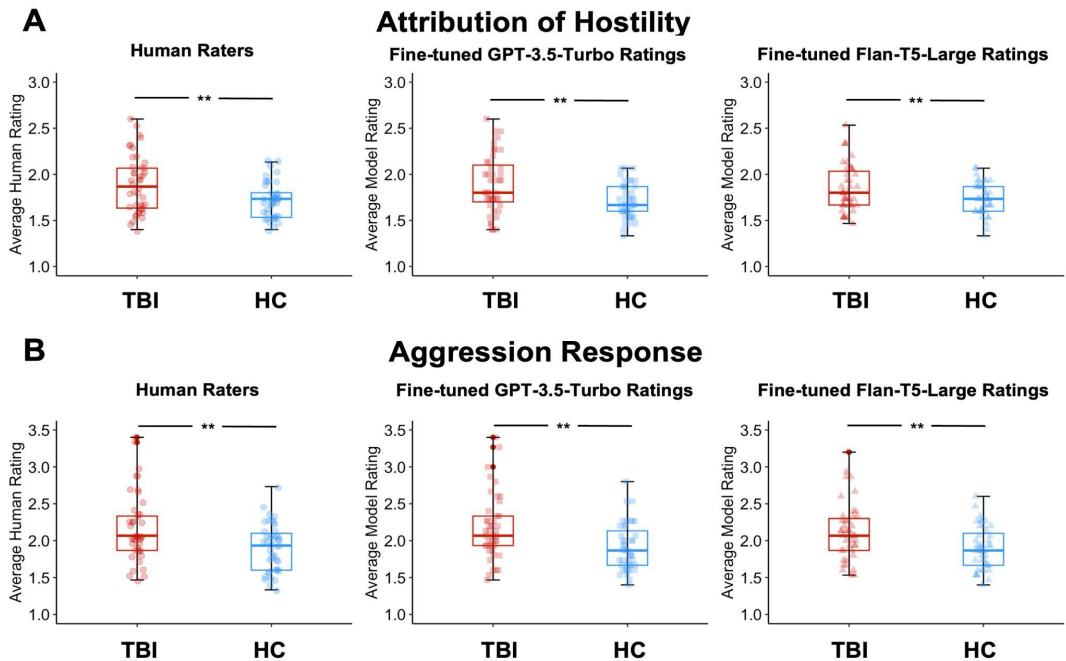

**Figure 3. The TBI group received significantly higher ratings than healthy controls when rated by human raters and the two fine-tuned language models. A.** *Attribution of hostility:* In the left panel, each datapoint represents a participant's mean hostility score averaged across two human raters (left), generated by fine-tuned GPT-3.5-Turbo (center) and generated by fine-tuned Flan-T5-Large (right). The central line of the boxplot denotes the median, the box bounds indicate the interquartile range, and the whiskers indicate the range. **B.** *Aggression response:* panels are organized as in A but display *aggression response* ratings. TBI and HC participants are displayed in red and blue respectively. ** indicates $p < 0.01$.

**Model-generated ratings reproduce relationships with self-report measures**

Another test of validity is whether model-generated ratings preserve established relationships among AIHQ subscales. Consistent with Neumann et al. (2020), *attribution of hostility* scores generated by the fine-tuned GPT-3.5-Turbo model were not significantly correlated with self-reported anger responses when averaged across all scenarios, but showed a significant positive correlation within ambiguous scenarios (Table 6). Neumann et al. (2020) also reported significant correlations between *aggression response* ratings and both attribution of intent and attribution of blame, across all scenarios and within ambiguous scenarios. These relationships were reproduced by the fine-tuned GPT-3.5-Turbo model, with aggression response ratings significantly correlated with both attribution subscales in both contexts (Table 6). The fine-tuned Flan-T5-Large model showed a similar pattern of associations across all analyses (Table A3), further supporting the validity of model-generated scores.

**Table 6.** *Correlations of AIHQ intent, attribution of hostility, and blame scales with AIHQ anger and aggression scales for TBI group (n = 85)*

| | Fine-tuned GPT-3.5-Turbo | | Trained human raters | |
|---|---|---|---|---|
| | **Attributions of hostility** $r$ | **Aggression response** $r$ | **Attributions of hostility** $r$ | **Aggression response** $r$ |
| **Attributions of intent** | | | | |
| All scenarios | 0.723*** | 0.485*** | 0.637*** | 0.507*** |
| Ambiguous scenarios | 0.763 *** | 0.426** | 0.684*** | 0.389* |
| Intentional Scenarios | 0.256 | 0.231 | 0.173 | 0.198 |
| Accidental scenarios | 0.755 *** | 0.482** | 0.783*** | 0.429** |
| **AIHQ anger response** | | | | |
| All scenarios | 0.247 | 0.674*** | 0.230 | 0.663*** |
| Ambiguous scenarios | 0.416** | 0.521*** | 0.474** | 0.506*** |
| Intentional Scenarios | 0.108 | 0.593*** | 0.079 | 0.573*** |
| Accidental scenarios | 0.300 | 0.723*** | 0.295 | 0.658*** |
| **Attributions of blame** | | | | |
| All scenarios | 0.544*** | 0.450** | 0.504*** | 0.459** |
| Ambiguous scenarios | 0.712*** | 0.348* | 0.695*** | 0.368* |
| Intentional Scenarios | 0.217 | 0.437** | 0.227 | 0.434** |
| Accidental scenarios | 0.542*** | 0.474** | 0.499*** | 0.428** |

*Note.* * $p < 0.05$. ** $p < 0.01$, *** $p < 0.001$

**Fine-tuned models generalize to a new dataset with high agreement with human ratings**

To assess the generalizability of the fine-tuned GPT-3.5-Turbo and Flan-T5-Large models, we applied them to a second, independent AIHQ dataset (Combs et al., 2013) that was not used during training. Model-generated ratings for attribution of hostility and aggression response were compared to scores provided by two trained human raters. Inter-rater reliability among the human raters was high for both *attribution of hostility* ($r = 0.896$, $t(142) = 23.99$, $p < .001$) and *aggression response* ratings ($r = 0.918$, $t(142) = 27.49$, $p < .001$).

Both fine-tuned models showed strong agreement with human ratings. Correlations exceeded 0.75 for *attribution of hostility* and 0.80 for *aggression response* (Figure 4). The pre-trained GPT-3.5-Turbo model performed moderately well without fine-tuning (*attribution of hostility*: $r = 0.656$, $t(142) = 10.35$, $p < .001$; *aggression response*: $r = 0.774$, $t(142) = 14.57$, $p < .001$), whereas the pre-trained Flan-T5-Large showed lower alignment for *attribution of hostility* ($r = 0.284$, $t(142) = 3.53$, $p < .001$) but moderately well alignment for *aggression response* ($r = 0.647$, $t(142) = 10.12$, $p < .001$). These results indicate that fine-tuning on the initial dataset substantially improves agreement with human scoring, even when applied to an unseen dataset.

We focused subsequent analyses on the fine-tuned models to assess performance across ambiguous, intentional, and accidental scenario types. Model-generated ratings and human ratings for *attribution of hostility* were highly correlated across ambiguous (fine-tuned GPT-3.5-Turbo: $r = 0.750$, $t(142) = 13.52$, $p < .001$; fine-tuned Flan-T5-Large: $r = 0.704$, $t(142) = 11.80$, $p < .001$), intentional (fine-tuned GPT-3.5-Turbo: $r = 0.770$, $t(142) = 14.39$, $p < .001$; fine-tuned Flan-T5-Large: $r = 0.735$, $t(142) = 12.91$, $p < .001$), and accidental (fine-tuned GPT-3.5-Turbo: $r = 0.705$, $t(142) = 11.85$, $p < .001$; fine-tuned Flan-T5-Large: $r = 0.762$, $t(142) = 14.02$, $p < .001$) scenarios. Inter-rater reliability among the two human coders also showed similar correlations across all three scenario types (ambiguous: r = 0.874, $t(142)$ = 21.42, $p < .001$; intentional: $r$ = 0.895, $t(142)$ = 23.86, $p < .001$; accidental: $r$ = 0.802, $t(142)$ = 16.03, $p < .001$).

Correlations between model-generated and human ratings for *aggression response* were lower for ambiguous scenarios (fine-tuned GPT-3.5-Turbo: $r$ = 0.539, $t(142)$ = 7.62, $p < .001$; fine-tuned Flan-T5-Large: $r$ = 0.480, $t(142)$ = 6.52, $p < .001$) compared to intentional (fine-tuned GPT-3.5-Turbo: $r$ = 0.917, $t(142)$ = 23.11, $p < .001$; fine-tuned Flan-T5-Large: $r$ = 0.844, $t(142)$ = 18.77, $p < .001$) and accidental scenarios (fine-tuned GPT-3.5-Turbo: $r$ = 0.889, $t(142)$ = 27.33, $p < .001$; fine-tuned Flan-T5-Large: $r$ = 0.847, $t(142)$ = 18.95, $p < .001$). A similar decrease was observed in inter-rater reliability among human coders (ambiguous: r = 0.775, $t(142)$ = 14.61, $p < .001$; intentional: $r$ = 0.935, $t(142)$ = 31.34, $p < .001$; accidental: $r$ = 0.923, $t(142)$ = 28.58, $p < .001$), suggesting that the lower correlations may reflect greater interpretive variability in responses to ambiguous scenarios in this dataset rather than a limitation of the models.

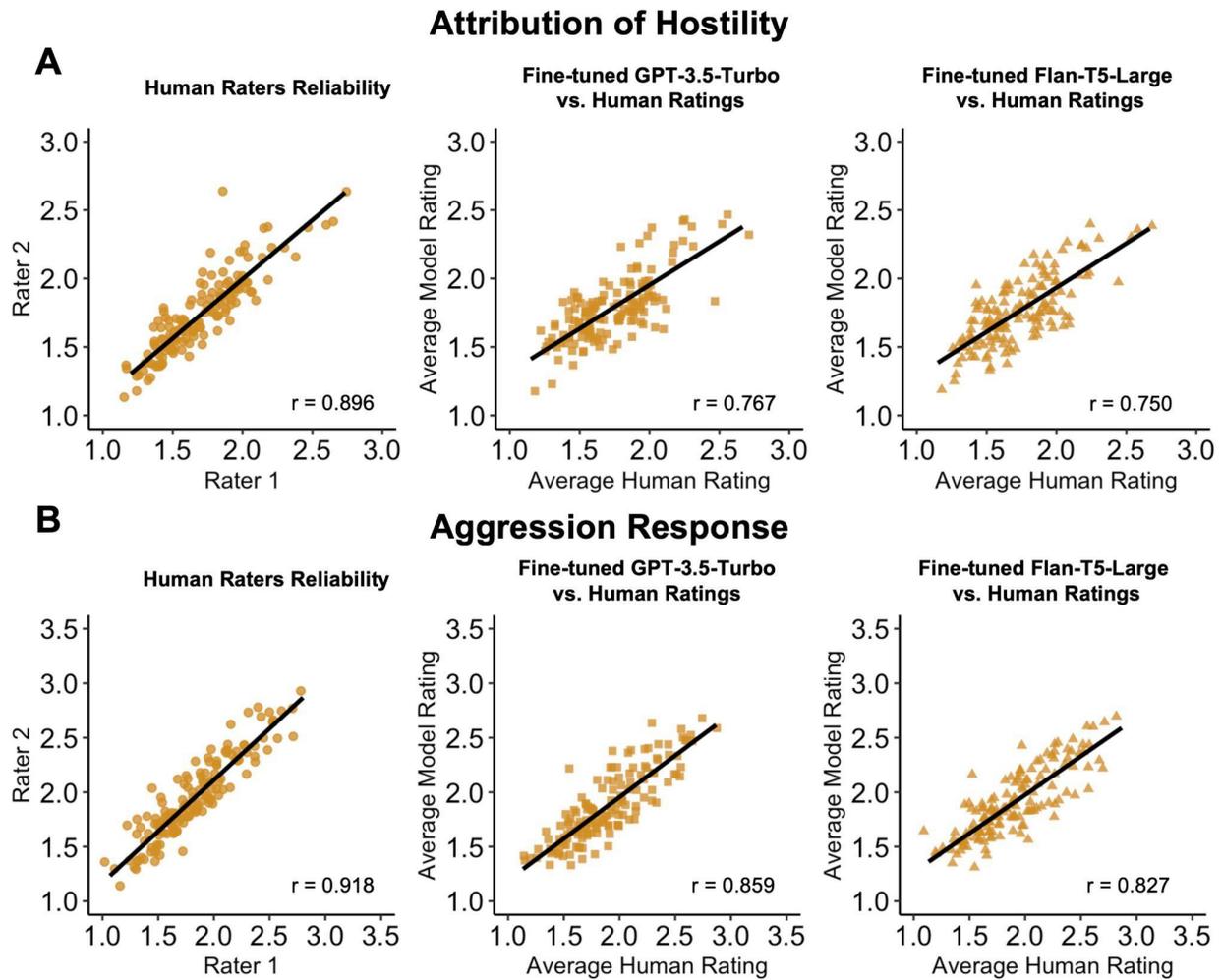

**Figure 4. Ratings given by fine-tuned GPT-3.5-Turbo and Flan-T5-Large highly correlated with human ratings on the second dataset. A.** *Attribution of hostility:* Each dot in the left panel represents each participant's rating for *attribution of hostility* from two human raters. The middle panel compares each participant's average human raters' rating with fine-tuned GPT-3.5-Turbo's rating, and the right panel compares each participant's average human raters' rating with fine-tuned Flan-T5-Large's rating. **B.** *Aggression response:* Same structure as A, but for *aggression response* ratings. Correlation coefficients *r* indicate how correlated the two ratings are in each panel.

**Validation in out-of-sample dataset**

We again compared correlations between *attribution of hostility*, *aggression response*, and other AIHQ items (attribution of intent, attribution of blame, anger response) across both models' ratings and human ratings in the second dataset. Fine-tuned GPT-3.5-Turbo and Flan-T5-Large ratings showed that *attribution of hostility* significantly correlated with attribution of intent across all AIHQ scenarios (ambiguous, intentional, and accidental), ambiguous scenarios only, and intentional scenarios only (Table 7, Table A4). *Attribution of hostility* was similarly correlated with attribution of blame for these same scenario categories. Using ratings given by the trained human raters, we found similar results on the significant correlations between *attribution of hostility* and attributions of intent and attributions of blame, with significant correlations across all scenarios, accidental only, ambiguous only, and intentional only (Table 7).

For *aggression response*, both GPT-3.5-Turbo and human raters' ratings indicated that *aggression response* was significantly correlated with attributions of intent when analyzing the full set of scenarios and the accidental-only subset (Table 7). Similar patterns of results are found using ratings by the fine-tuned Flan-T5-Large (Table A4). *Aggression response* also significantly correlated with the anger response and attributions of blame across all scenario types, when rated both by fine-tuned GPT-3.5-Turbo and human raters (Table 7). These findings suggest stable relationships between AIHQ measures regardless of whether they were rated by human coders or the two language models.

**Table 7.** *Correlations of AIHQ intent, attribution of hostility, and blame scales with AIHQ anger and aggression scales on the new dataset.*

| | Fine-tuned GPT-3.5-Turbo | | Trained human raters | |
|---|---|---|---|---|
| | **Attributions of hostility r (*p* value)** | **Aggression response r (*p* value)** | **Attributions of hostility r (*p* value)** | **Aggression response r (*p* value)** |
| **Attributions of intent** | | | | |
| All scenarios | 0.485*** | 0.323*** | 0.486*** | 0.311*** |
| Ambiguous scenarios | 0.616*** | 0.227** | 0.594*** | 0.178* |
| Intentional Scenarios | 0.314*** | 0.234** | 0.304*** | 0.242** |
| Accidental scenarios | 0.249** | 0.384*** | 0.344*** | 0.374*** |
| **AIHQ anger response** | | | | |
| All scenarios | 0.426*** | 0.448*** | 0.408*** | 0.492*** |
| Ambiguous scenarios | 0.466*** | 0.396*** | 0.459*** | 0.358*** |
| Intentional Scenarios | 0.373*** | 0.428*** | 0.386*** | 0.462*** |
| Accidental scenarios | 0.303*** | 0.516*** | 0.363*** | 0.532*** |
| **Attributions of blame** | | | | |
| All scenarios | 0.396*** | 0.453*** | 0.379*** | 0.464*** |
| Ambiguous scenarios | 0.502*** | 0.322*** | 0.462*** | 0.260*** |
| Intentional Scenarios | 0.352*** | 0.388*** | 0.334*** | 0.381*** |
| Accidental scenarios | 0.206* | 0.486*** | 0.304*** | 0.521*** |

*Note.* * $p < 0.05$. ** $p < 0.01$, *** $p < 0.001$

## Discussion

We fine-tuned and validated two large language models, GPT-3.5-Turbo and Flan-T5-Large, to generate automated ratings for responses to two open-ended questions from the Ambiguous Intentions Hostility Questionnaire (AIHQ). Both models demonstrated a high degree of consistency with ratings provided by human raters, supporting their reliability in assessing the *attribution of hostility* and *aggression response* across participant groups and scenario types. Fine-tuned models produced ratings that more closely aligned with human judgments, achieving correlation values comparable to, and sometimes surpassing, those between two independent human raters. Importantly, the models also reproduced previously reported group differences between individuals with traumatic brain injury and healthy controls, indicating sensitivity to clinically meaningful variability. Moreover, when tested on a new dataset not included in training, the fine-tuned models again showed strong correlations with human ratings. This demonstrates the models' ability to generalize and accurately score novel AIHQ responses. Together, these findings suggest that large language models offer a viable and efficient method for automating the scoring of open-ended psychological assessments, substantially reducing the need for time-intensive manual rating.

The Ambiguous Intentions Hostility Questionnaire (AIHQ) is an important tool for measuring hostile attribution bias, capturing how individuals interpret ambiguous social interactions—a bias that is closely tied to aggression, interpersonal conflict, and mental health challenges (Dodge et al., 2015; Pettit et al., 2010; H. L. Smith et al., 2016). The AIHQ's open-ended questions allow participants to freely express their interpretations, capturing nuanced biases that may be missed by Likert-scale measures (Barbara A. Woike, 2009). However, traditional scoring of these open-ended responses is time-intensive, requiring trained raters to spend hours evaluating each participant's responses. This burden limits the scalability of the AIHQ in large-sample or longitudinal studies, and introduces potential variability due to differences in rater interpretation. By leveraging large language models to automate AIHQ scoring, researchers can dramatically reduce the time and labor involved in data processing, while improving the consistency and reproducibility of ratings across study contexts. This approach enables broader application of the AIHQ in both research and clinical contexts.

Model-generated ratings for attribution of hostility and aggression response in the AIHQ showed significant correlations with related AIHQ subscales, including attribution of intent, anger, and blame, providing evidence for the convergent validity of the automated rating method. These correlations closely mirrored those obtained from human raters (Combs et al., 2013; Neumann et al., 2020), reinforcing the model's ability to capture core dimensions of hostile attribution bias. This convergent validity suggests that model-generated scores align not just statistically with human ratings, but also conceptually with the psychological constructs being measured. The ability of large language models to replicate these established relationships strengthens confidence in their use as scalable tools for assessing hostile attribution bias in both research and applied settings.

A meta-analytic review of 41 studies found a robust association between hostile attribution of intent and aggressive behavior, particularly in ecologically valid contexts (De Castro et al., 2002). Increased awareness of hostile attribution bias among individuals with aggressive tendencies can open the door to targeted strategies aimed at improving interpretations of ambiguous social situations (Klein Tuente et al., 2019). Studies have shown that interventions targeting hostile attribution bias are effective in reducing reactive aggression, offering promising pathways for aggression regulation training (Guerra & Slaby, 1990; Neumann et al., 2023). By providing a reliable and efficient method for assessing hostility-related biases, large language models offer scalable tools to support intervention development. Model-generated ratings could be used not only for screening but also to monitor change in attributional style over time and to evaluate intervention effectiveness.

Hostile attribution bias is also implicated in a range of psychiatric conditions, including schizophrenia, depression, and anxiety (Buck et al., 2023; Eysenck et al., 1991; Mogg et al., 2006; H. L. Smith et al., 2016) and individuals with TBI (Neumann et al., 2017, 2020; Neumann, Sander, Perkins, et al., 2021; Neumann, Sander, Witwer, et al., 2021). The tendency to interpret ambiguous social situations in a hostile manner can reinforce maladaptive thought patterns and contribute to symptom persistence. When paired with realistic vignettes, the AIHQ offers an ecologically valid window into how clinical patients differ in how they interpret social intent, which may not be as easily captured by traditional self-report measures. Although the AIHQ has been used to study HAB in clinical populations, its widespread adoption is limited by the demands of manual scoring. Automated scoring with large language models overcomes this barrier, enabling broader clinical use of the AIHQ while preserving the open-ended, naturalistic responses.

Despite the strong performance of the fine-tuned models, several limitations should be noted. First, the models were fine-tuned on ratings provided by a specific group of human raters, and their performance may vary depending on scoring conventions and cultural context. Indeed, when we applied the fine-tuned models to a new undergraduate dataset rated by a different group of coders, model-human correlations were approximately 0.10 lower than those observed in the original held-out test set, suggesting that differences in rater training or participant demographics can reduce alignment. Cross-cultural research has also shown that both response styles and patterns of hostile attribution differ across societies (Zajenkowska et al., 2021). For example, Zajenkowska and colleagues (2021) found that Polish participants attributed less hostility to close friends, U.S.

participants showed higher hostility toward acquaintances and strangers, and Japanese participants showed heightened hostility in interactions with new co-workers. The AIHQ includes scenarios involving a range of relationship types (e.g., friend, stranger, co-worker, authority figure), and models fine-tuned on U.S.-based data may internalize patterns of social interpretation that reflect American norms and rater expectations. Future work should explore fine-tuning on larger, more culturally diverse datasets to enhance generalizability and ensure that automated scoring tools remain valid across varied populations and settings.

Another limitation is that, in our current approach, open-ended AIHQ responses are ultimately reduced to a single numerical rating per construct. While this aligns with existing scoring conventions, it may underutilize the depth and nuance present in participants' natural language explanations. Recent work suggests that large language models can extract a broader range of psychologically meaningful features from open-ended text, including affective tone and well-being indicators (Kjell et al., 2024; Nilsson et al., 2024; Tanana et al., 2021). Future research could investigate whether generating more detailed and multidimensional outputs, such as differentiating between proactive and reactive forms of aggression (Dodge & Coie, 1987), or identifying specific emotional or cognitive themes, may enable more comprehensive psychological assessment and enhance the theoretical utility of the AIHQ.

**Practical considerations**

While both GPT-3.5-Turbo and Flan-T5-Large performed well in generating automated AIHQ ratings, each model presents distinct trade-offs in terms of usability, accessibility, and data privacy. In particular, GPT-3.5-Turbo is accessed through the OpenAI API, is fast and involves minimal setup. However, due to OpenAI's current policy, the fine-tuned version of the model cannot be publicly shared. In our study, the base GPT-3.5-Turbo model, when used without fine-tuning, performed slightly worse than the fine-tuned version but still achieved acceptable correlations with human ratings, making it a reasonable alternative for many use cases. Use of the OpenAI API incurs a fee, and a more significant limitation is that input data must be transmitted to OpenAI's servers, which may not be compatible with institutional restrictions around data sharing.

In contrast, Flan-T5-Large is a fully open-source model that can be downloaded and run locally, either on a personal computer or a secure computing cluster. While this requires more initial setup, it offers full control over the model environment and preserves data privacy by eliminating the need to send responses to an external server. Importantly, the fine-tuned version of Flan-T5-Large developed in this study, which shows superior performance, can be shared and reused by other researchers without restriction. Thus, GPT-3.5-Turbo is better suited for large, less sensitive datasets, while the fine-tuned Flan-T5-Large model is better suited for smaller datasets that involve sensitive or protected data.

To support broader adoption, we developed a Google Colab notebook that allows researchers to run the Flan-T5-Large model or the base GPT-3.5-Turbo model without requiring local installation. This option offers an accessible cloud-based environment that can be used directly in a web browser, though it transmits data to Google servers for computation. For users who prefer a local solution, we also provide a drag-and-drop browser-based interface that allows researchers to upload a CSV file of participants' AIHQ free-text responses and generate ratings for both attribution of hostility and aggression responses. The tool supports both the fine-tuned Flan-T5-Large model and the base GPT-3.5-Turbo model (for users with an OpenAI API key). When using the Flan-T5-Large model locally, no data is transmitted to external servers, providing the     most     secure     option.     Full     installation     instructions     are     available     at: https://aihqrating.readthedocs.io.

**Conclusion**

In this study, our findings demonstrate that large language models can reliably automate the scoring of open-ended responses in the Ambiguous Intentions Hostility Questionnaire, closely replicating human judgment across participant groups, scenario types, and datasets. By retaining the nuance of open-ended responses while removing the need for time-intensive manual scoring, our approach expands the feasibility of using the AIHQ in both large-scale and clinically sensitive research. Moreover, the availability of both commercial and open-source model implementations offers flexibility for a range of use cases, balancing scalability, data privacy, and accessibility. As natural language models continue to improve, their integration into psychological research offers a promising path for increasing the scalability of assessment and advancing the study of social cognition in mental health.

https://doi.org/10.1080/00223891.2020.1774380

<div align="center">**Additional Information**</div>

**Model Prompting and Fine-Tuning**

We set the system prompt to be the default for GPT-3.5-Turbo: "You are a helpful assistant." For in-context learning, we prompted both GPT-3.5-Turbo and Flan-T5-large models to give numerical ratings from 1 to 5 for participants' hostility and aggression responses to AIHQ scenarios:

> "Read the following written responses people give regarding why they thought the person in the scene before acted toward them the way they did. Your task is to rate the hostility of the attributed intent. A rating of 1 would be given if someone perceived the situation as an accident. A rating of 5 would be given if someone thought the person in the scene did this on purpose and wanted to harm them. A rating of 3 might be given if the person thought the person in the scene did this on purpose to some degree, but did not intend to harm them. Please output a single numeric number from 1 to 5, do not include anything other than the number in the output."

> "Read the following behavioral responses people give toward the social situation above. Your task is to rate the presence of aggression in the behavioral response. A rating of 1 is given for a passive response or one in which the participant says that she/he would do nothing. A rating of 5 would be given for physical retaliation. Other examples of ratings include: 2 (the participant reports that he /she would ask why the other person acted toward them in that way), 3 (the participant would tell the other person not to act that way again), and 4 (the participant would yell at the other person). Please output a single numeric number from 1 to 5, do not include anything other than the number in the output."

To further improve the models' performance, we randomly subsampled half of the TBI participants (n = 42) and half of the HC participants (n = 42), and used their responses to all 15 AIHQ scenarios as training data (Figure 1). The training dataset is reformatted into the conversational chat format to fine tune GPT-3.5-Turbo and the user-question format to fine tune Flan-T5-Large. We then give the same prompt to the two fine-tuned models to rate the hostility and aggression responses to AIHQ scenarios given by participants not in the training dataset.

Fine-tuning GPT-3.5-Turbo aimed to enhance its ability to rate responses in a way that closely aligns with trained human raters. Following OpenAI's fine-tuning guidelines, the model was trained using a conversational chat format consistent with its actual application. The training examples incorporated the exact prompts the model would encounter in the rating task, with the responses structured as single numerical ratings. This approach helped the model learn to accurately identify and differentiate varying levels of hostility and aggression from the training data, refining its ability to provide appropriate ratings. By aligning the training data with the intended use case, the fine-tuned GPT-3.5-Turbo demonstrated improved proficiency in producing ratings that resemble those given by human raters.

**Table A1.** *Correlation between the ratings given by human raters and ratings given by fine-tuned Flan-T5-Large on half of TBI group data and half of HC group data.*

| Variable | Fine-tuned Flan-T5-Large | | Inter-Rater Correlation | |
|---|---|---|---|---|
| | **TBI Group (n = 43)** | **HC Group (n = 43)** | **TBI Group (n = 43)** | **HC Group (n = 43)** |
| **Attributions of hostility** | | | | |
| All scenarios | 0.897 | 0.795 | 0.897 | 0.801 |
| Ambiguous scenarios | 0.849 | 0.806 | 0.916 | 0.802 |
| Intentional Scenarios | 0.782 | 0.814 | 0.758 | 0.730 |
| Accidental scenarios | 0.768 | 0.515 | 0.936 | 0.716 |
| **AIHQ aggression response** | | | | |
| All scenarios | 0.944 | 0.864 | 0.940 | 0.906 |
| Ambiguous scenarios | 0.906 | 0.737 | 0.908 | 0.726 |
| Intentional Scenarios | 0.894 | 0.851 | 0.914 | 0.921 |
| Accidental scenarios | 0.868 | 0.832 | 0.912 | 0.881 |

*Note.* All correlations were significant at $p < .001$.

**Table A2.** *Testing for the differences in the means of 2 groups for attribution of hostility and aggression response.*

| Variable | Human rated | | | Fine-tuned Flan-T5-Large rated | | |
|---|---|---|---|---|---|---|
| | TBI (n=43) | HC (n=43) | TBI > HC $p$ value | TBI (n = 43) | HC (n = 43) | TBI > HC $p$ value |
| **Attributions of hostility** | | | | | | |
| All scenarios | 1.90 | 1.72 | .005** | 1.84 | 1.73 | .039* |
| Ambiguous scenarios | 1.97 | 1.69 | .017* | 1.97 | 1.72 | .015* |
| Intentional Scenarios | 2.42 | 2.35 | .357 | 2.42 | 2.38 | .958 |
| Accidental scenarios | 1.29 | 1.11 | .005** | 1.29 | 1.11 | .115 |
| **AIHQ aggression response** | | | | | | |
| All scenarios | 2.17 | 1.91 | .005** | 2.12 | 1.90 | .006** |
| Ambiguous scenarios | 2.07 | 1.84 | <.001*** | 2.04 | 1.82 | < .001*** |
| Intentional Scenarios | 2.45 | 2.21 | .064 | 2.42 | 2.31 | .307 |
| Accidental scenarios | 1.99 | 1.68 | .023* | 1.91 | 1.58 | .006** |

*Note.* Average ratings given by large language models trained on 50% of data are significantly higher for the TBI compared to the HC group, reproducing results obtained with human raters. * $p < 0.05$. ** $p < 0.01$, *** $p < 0.001$

**Table A3.** *Correlations of AIHQ intent, attribution of hostility, and blame scales with AIHQ anger and aggression scales for TBI group (n = 85)*

| | Fine-tuned Flan-T5-Large | | Trained human raters | |
|---|---|---|---|---|
| | **Attributions of hostility r (*p* value)** | **Aggression response r (*p* value)** | **Attributions of hostility r (*p* value)** | **Aggression response r (*p* value)** |
| **Attributions of intent** | | | | |
| All scenarios | 0.503*** | 0.503*** | 0.637*** | 0.507*** |
| Ambiguous scenarios | 0.597*** | 0.362* | 0.684*** | 0.389* |
| Intentional Scenarios | 0.105 | 0.212 | 0.173 | 0.198 |
| Accidental scenarios | 0.539*** | 0.30* | 0.783*** | 0.429** |
| **AIHQ anger response** | | | | |
| All scenarios | 0.340* | 0.659*** | 0.230 | 0.663*** |
| Ambiguous scenarios | 0.433** | 0.430** | 0.474** | 0.506*** |
| Intentional Scenarios | 0.143 | 0.545*** | 0.079 | 0.573*** |
| Accidental scenarios | 0.320* | 0.585*** | 0.295 | 0.658*** |
| **Attributions of blame** | | | | |
| All scenarios | 0.544*** | 0.463** | 0.504*** | 0.459** |
| Ambiguous scenarios | 0.625*** | 0.301 | 0.695*** | 0.368* |
| Intentional Scenarios | 0.184 | 0.469** | 0.227 | 0.434** |
| Accidental scenarios | 0.504*** | 0.307* | 0.499*** | 0.428** |

*Note.* * $p < 0.05$. ** $p < 0.01$, *** $p < 0.001$

**Table A4.** *Correlations of AIHQ intent, attribution of hostility, and blame scales with AIHQ anger and aggression scales on the new dataset.*

| | Fine-tuned Flan-T5-Large | | Trained human raters | |
|---|---|---|---|---|
| | **Attributions of hostility r (*p* value)** | **Aggression response r (*p* value)** | **Attributions of hostility r (*p* value)** | **Aggression response r (*p* value)** |
| **Attributions of intent** | | | | |
| All scenarios | 0.456*** | 0.311*** | 0.486*** | 0.311*** |
| Ambiguous scenarios | 0.593*** | 0.324*** | 0.594*** | 0.178* |
| Intentional Scenarios | 0.233** | 0.212* | 0.304*** | 0.242** |
| Accidental scenarios | 0.321*** | 0.330*** | 0.344*** | 0.374*** |
| **AIHQ anger response** | | | | |
| All scenarios | 0.399*** | 0.448*** | 0.408*** | 0.492*** |
| Ambiguous scenarios | 0.496*** | 0.429*** | 0.459*** | 0.358*** |
| Intentional Scenarios | 0.275*** | 0.411*** | 0.386*** | 0.462*** |
| Accidental scenarios | 0.308*** | 0.437*** | 0.363*** | 0.532*** |
| **Attributions of blame** | | | | |
| All scenarios | 0.371*** | 0.411*** | 0.379*** | 0.464*** |
| Ambiguous scenarios | 0.506*** | 0.344*** | 0.462*** | 0.260*** |
| Intentional Scenarios | 0.276*** | 0.341*** | 0.334*** | 0.381*** |
| Accidental scenarios | 0.223** | 0.436*** | 0.304*** | 0.521*** |

*Note.* * $p < 0.05$. ** $p < 0.01$, *** $p < 0.001$